\documentclass[letterpaper, 10 pt, conference]{ieeeconf}  % Comment this line out if you need a4paper

\IEEEoverridecommandlockouts                              % This command is only needed if 
\usepackage{graphics} % for pdf, bitmapped graphics files
\usepackage{epsfig} % for postscript graphics files
\usepackage{mathptmx} % assumes new font selection scheme installed
\usepackage{times} % assumes new font selection scheme installed
\usepackage{amsmath} % assumes amsmath package installed
\usepackage{amssymb}  % assumes amsmath package installed

\usepackage{fancyhdr}
\fancypagestyle{firststyle} 
{
   \fancyhf{}
   \fancyfoot[C]{\scriptsize © 2021 IEEE.  Personal use of this material is permitted. Permission from IEEE must be obtained for all other uses, in any current or future media, including reprinting/republishing this material for advertising or promotional purposes, creating new collective works, for resale or redistribution to servers or lists, or reuse of any copyrighted component of this work in other works.}
}

\title{\LARGE \bf
Improving the compromise between accuracy, interpretability and personalization of rule-based machine learning in medical problems 
}

\author{Francisco Valente$^{1}$, Jorge Henriques$^{1}$, Simão Paredes$^{2}$, Teresa Rocha$^{2}$, Paulo de Carvalho$^{1}$, João Morais$^{3}$% <-this % stops a space

\thanks{$^{1}$F. Valente, J. Henriques and P. de Carvalho are with the Centre for Informatics and Systems of the University of Coimbra
        {\tt\small pfcv@dei.uc.pt}}%
\thanks{$^{2}$S. Paredes and T. Rocha are with the Polytechnic of Coimbra, Coimbra Institute of Engineering}
\thanks{$^{3}$J. Morais is with the Cardiology Department of Leiria Hospital Centre}%
}

\begin{document}

\maketitle
\thispagestyle{firststyle}

%%%%%%%%%%%%%%%%%%%%%%%%%%%%%%%%%%%%%%%%%%%%%%%%%%%%%%%%%%%%%%%%%%%%
\begin{abstract}

One of the key challenges when developing a predictive model is the capability to describe the domain knowledge and the cause-effect relationships in a simple way. Decision rules are a useful and important methodology in this context, justifying their application in several areas, particularly in clinical practice. Several machine-learning classifiers have exploited the advantageous properties of decision rules to build intelligent prediction models, namely decision trees and ensembles of trees (ETs). However, such methodologies usually suffer from a trade-off between interpretability and predictive performance. Some procedures consider a simplification of ETs, using heuristic approaches to select an optimal reduced set of decision rules. In this paper, we introduce a novel step to those methodologies. We create a new component to predict if a given rule will be correct or not for a particular patient, which introduces personalization into the procedure. Furthermore, the validation results using three public clinical datasets suggest that it also allows to increase the predictive performance of the selected set of rules, improving the mentioned trade-off.
\end{abstract}

%%%%%%%%%%%%%%%%%%%%%%%%%%%%%%%%%%%%%%%%%%%%%%%%%%%%%%%%%%%%%%%%%%%%%%%%%%%%%%%%
\section{INTRODUCTION}

Physicians usually incorporate clinical decision rules (CDRs) in several domains of their practice, such as bedside diagnostic or therapeutic choices. CDRs appeared as a tool to help the decision-making of the clinical staff, reducing its uncertainty and making it more evidence-based \cite{Stiell2007}. In the last few decades, several machine learning (ML) methods have been proposed as decision support systems in almost every medical specialties, where they have been shown to achieve high performances \cite{Topol2019}. However, the majority of those models are often seen as a “black-box” with a lack of explainability capabilities, which limits their use in medicine. In fact, despite a large investment in the development of novel ML applications in medical areas, its translation to the daily clinical practice is still very limited \cite{Mateen2020}.
\par
Decision trees (DTs) are widely acknowledged as a very interpretable ML approach, and therefore a useful solution when the scenario requires a deep understanding of the generated model \cite{SAGI2020124}, such as the medical cases. DTs present several features that make them appealing from the interpretability point of view, such as: 1) they mimic the human reasoning, incorporating a combination of decision rules; 2) the IF-THEN nature of such rules allow to easily extract domain knowledge from the cause-effect relationships; 3) the tree-like visual representation makes the overall classification process easy to understand. Despite these characteristics, DTs typically present a worse predictive performance when compared to more complex ML methodologies. 
\par
In order to overcome that disadvantage, ensembles of trees (ET) were proposed. ET are based on the idea that the combination of several weaker classifiers (several DTs) achieves better performance than a single classifier (one DT). Random Forest (RF) \cite{Breiman2001}, a tree ensemble methodology, is one of the most widely applied classifiers as it often assures a high performance \cite{Delgado2014}. In RF, multiple DTs are built, each one using a randomly selected subset of the whole set of features, and (optionally) also a subset of the whole set of samples. The outputs of the individual trees are then combined to produce the final output. A DT is easy to interpret individually, as long as its dimensions (number of rules and length of each rule) keeps low. However, to study several (hundreds or thousands) DTs simultaneously is unfeasible for the final user (e.g. physicians). Thus, RFs are typically considered as “black-box” models. 
\par
Therefore, there is a trade-off between interpretability and accuracy in such rule-based ML methodologies. In order to overcome it, some approaches have been proposed, mainly suggesting techniques to improve the interpretability of the generated ET \cite{SAGI2020124}\cite{Moore2018}\cite{Hara2018}. A  group of procedures aims at simplifying the ET by decomposing the respective several individual DTs into a set of decision rules. Then, they select a sparser set of the best rules, based on heuristics such as LASSO \cite{Friedman2008}, hill climbing \cite{Mashayekhi2017} or quadratic programs \cite{Meinshausen2010}. These approaches allow to obtain a set of important decision rules directly extracted from the ET, which is simpler to analyze than a group of DTs, improving the interpretability. Even so, those methods often select a final set of several dozens of rules, which is still a high amount of rules for an easy and fast interpretation of the output and the respective extracted knowledge in the clinical practice.  Furthermore, the final set of rules is usually applied uniformly to all patients, i.e. the decision rules have the same weight to all samples.
\par
In this study, we propose an approach that predicts the correctness of each one of the decision rules to each patient, enabling to use a smaller set of decision rules to obtain the same performance of state-of-the-art methods. The goal is to promote a novel methodology towards a more interpretable and personalized set of decision rules, while keeping a good predictive ability, improving its usability in the clinical field.

\section{METHODS}

\subsection{Generation and extraction of decision rules}

A decision tree can be converted to a set of decision rules. An individual decision rule corresponds to a path from the roof of the tree to its leaf.  The condition of the rule, i.e. the IF segment, is defined 
by the set of conditions in that path. The prediction, i.e. the THEN segment, is defined by the class attributed to the leaf. Thus, there are as many rules as leaves in the tree. Furthermore, assuming a binary classification, we may consider that if the condition part of a given rule is not verified, then at that decision level the rule leads to a prediction of the opposite class (represented by an ELSE segment). Fig. 1 exemplifies a very simple DT. From that DT, the following decision rules can be extracted:

\begin{enumerate}
  \item IF x1 $>$ 0.5, THEN y1 (ELSE y2).
  \item IF x1 $\leq$ 0.5 AND x2 = false, THEN y2 (ELSE y1).
  \item IF x1 $\leq$ 0.5 AND x2 = true, THEN y1 (ELSE y2).
\end{enumerate}

\begin{figure}[thpb]
\centering
\includegraphics[width=0.3\textwidth]{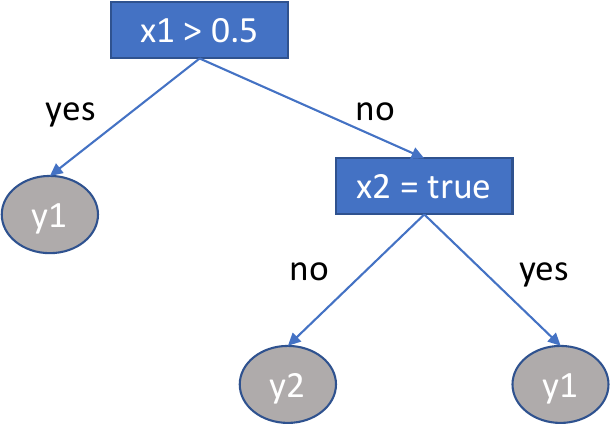}
\caption{Example of a decision tree. The split nodes are represented by the blue square boxes, and the leaf (or terminal) nodes are represented by the grey round boxes. $x1$ and $x2$ are two attributes (features), and $y1$ and $y2$ are two arbitrary classes (labels).}
\label{fig:tree}
\end{figure}

We take advantage of the random forest ability to generate a large group of DTs, and then get a set of decision rules as aforementioned. For interpretability issues, it is not enough to have decision rules, they must also be easy to understand, which implies those rules should be short. Therefore, in this study, we set to 3 the maximum depth of each DT, i.e. the condition part of each rule is composed of at most 3 elements (two AND connections).

\subsection{Selection of a set of the best decision rules}

In this study, the decision rules extracted from the individual trees of the RF are combined into a single set of rules. Then, we start by removing duplicated rules. Further, a subset of M decision rules is obtained using a logistic regression model with L1-regularization (LASSO), similar as performed in \cite{Friedman2008}, as LASSO is a simple and intelligible heuristic selection method. 
\par
In this LASSO step, the input is a NxP matrix, where N is the number of samples and P is the number of decision rules extracted from the RF, and the output is the Nx1 outcome vector (observed outputs). The NxP matrix is binary, with a value of 1 if the rule condition (IF-part) of the rule $p$ is verified to the patient $n$, and a value of 0 otherwise. Let us consider the rules of the DT of Fig. \ref{fig:tree}. If a given sample has the values $x1=0.3$ and $x2=true$, only the IF-part of rule 3 is verified. Thus, the input vector for such patient and rules will be $condition\_verified([r1\text{ }r2\textbf{ }r3])$ = [0 0 1]. This procedure is applied to the N patients, taking into account the P rules, creating then the NxP matrix.
\par
Therefore, the LASSO is applied as a rules selection procedure, shrinking towards 0 the coefficients of the rules that are least relevant to the output prediction as well as the ones of correlated rules. In our approach, we then obtain a subset of M (relevant and uncorrelated) rules by selecting the ones that present the M highest LASSO regression coefficient values. The M value can be defined by the user. For a better generalization ability, a 3-fold cross-validation LASSO is applied for that task.

\subsection{Prediction of the correctness of each rule}

In this study, we aim to classify each rule for each sample, i.e. to predict when a rule will give a correct output or not. In order to accomplish that goal, we create a classification model for each rule. More specifically, we train a model giving as features the variables used by all the selected rules, and as label a binary vector with a value of 1 if the rule gave a correct output and a value of 0 if the rule gave a wrong prediction. For example, considering the DT of Fig. 1, if a given sample has the values $x1=0.3$ and $x2=true$, the predictions of the rules will be: 

\begin{itemize}
    \item rule 1: $y2$ ; rule 2: $y1$ ; rule 3: $y1$
\end{itemize}

So, if we assume that the true output is $y1$, rule 1 is incorrect (label=0) and rules 2 and 3 are correct (label=1). Therefore, if e.g. 5 rules are selected (M=5), 5 different binary label vectors will be created (with the information about this correctness for all the samples for each rule). Consequently, 5 classifiers are trained, each one to predict the individual rule correctness. The features used to train such classifiers are the original risk factors used by that selected rules. Thus, if those 5 rules use a total of K features, each of the 5 classifiers will have as input the NxK predictors matrix. Any classification algorithm can be used to predict the rule's correctness. In this study, a LASSO model was considered.
\par
So, we not only extract and selected a subset of rules from an ET, as in state-of-the-art methods, but we also attempt to forecast if a given rule should be (or not) applied to a particular patient. This procedure introduces personalization to the methodology as the rules will be applied differently and more properly for each patient, which can contribute to the improvement of the individual predictions. Thus, this step is a novelty in relation to literature approaches, which apply the decision rules uniformly to all patients.

\subsection{Computation of the probability of each class}

For each new (validation) sample, two vectors of dimension 1xM are then generated, composed of binary values. The predicted rule's output informs if a given rule classifies the patient as having a disease (1) or not (0). The predicted rule's correctness informs if a given rule is expected to be correct (1) or not (0) for that patient. Table \ref{tab:vectors} presents an example of the vectors obtained for a given new sample, using 3 rules.

\begin{table}[htbp]
\centering
\caption{Predictions obtained for a new sample, using a set of 3 rules.}
\label{tab:vectors}
\begin{tabular}{|l|c|c|c|}
\hline
\textbf{}                                                             & \textbf{Rule 1} & \textbf{Rule 2} & \textbf{Rule 3} \\ \hline
\begin{tabular}[c]{@{}l@{}}Rule's output  prediction\end{tabular}     & 1               & 0              & 1              \\ \hline
\begin{tabular}[c]{@{}l@{}}Rule's correctness prediction\end{tabular} & 1        & 1      & 0       \\ \hline
\end{tabular}
\end{table}

In order to generate the final output (probability that the sample is a positive class - e.g., the probability the patient has a given disease), the standard methodologies only do an averaging of the set of the M selected decision rules, usually:

\begin{equation}
\displaystyle Probability (class=1) ={\frac {1}{M}}\sum _{i=1}^{M}rule\_output_{i}.
\end{equation}

For the sample exemplified in Table \ref{tab:vectors}, those approaches would give a positive class probability of $\sim$0.66, because 2 rules out of 3 predict a positive class.

In contrast, in our approach, we also take into account the information about if each rule is expected to be (or not) correct for a given patient, i.e. the information of the last row of Table \ref{tab:vectors}. More specifically, the probability that a sample belongs to the positive class is given by a weighted average, which is personalized for each patient:

\begin{equation}
\displaystyle Probability (class=1) = {\frac {\sum _{i=1}^{M}rule\_output_{i}\cdot weight_{i}}{\sum_{i=1}^{M}weight_{i}}},
\end{equation}

We could do only a simple averaging of the rules predicted to be correct, giving a weight of 1 if the rule is predicted to be correct, and a weight of 0 otherwise. However, it may happen that for some samples, none of the M selected rules is predicted to be correct, and thus no rule would be available to predict the final outcome. Therefore, we assume a weight of 2 if the rule is predicted to be correct, and a weight of 1 otherwise. This implies that the methodology can be generalized to all scenarios. Furthermore, it means that all the rules are considered but the ones predicted to be correct for that patient contribute twice more for the final output than the others. Therefore, for the example of Table \ref{tab:vectors}, rules 1 and 2 have a weight of 2, and rule 3 has a weight of 1. Thus, the probability of the positive class will be 0.6. Surely, such weights can be adjusted or optimized by the user.

\subsection{Validation of the proposed approach}

The proposed approach was validated in three public clinical datasets. Two of them are from UCI Machine Learning Repository (\textit{https://archive.ics.uci.edu}): Heart Disease (prediction of presence/absence of heart disease) and Breast Cancer Wisconsin Diagnostic (prediction of benign/malign diagnosis of breast cancer). The third one is from Kaggle (\textit{https://www.kaggle.com/}): Pima Indians Diabetes Database (prediction of presence/absence of diabetes disease). The datasets will be designated as Heart, Breast and Diabetes, respectively. Heart has some categorical variables with missing data, which was replaced by the most frequent value of the corresponding feature. A 10-times repeated 5-fold stratified cross-validation was used as the validation procedure.

\section{RESULTS}

The ability of the proposed approach to correctly classify the data into positive (presence of disease) or negative (absence of disease) samples was assessed through the area under the ROC curve (AUC). Those results are presented in Table \ref{tab:auc}. The results of the proposed methodology are presented for different sets of selected rules, i.e. for different M values. More specifically, sets with the 3, 5, 10, 15 and 20 best decision rules were considered, as selected by the LASSO approach. The initial set of P rules was obtained by building a RF with 100 DTs. The results are also compared with two standard rules-based machine learning models: random forest and decision tree. For the RF methodologies, two versions are considered: one where the RF and its trees can grow without any constrictions; and a simpler and more interpretable version, where the RF can have at most 5 trees, each one with a maximum depth of 3. The parameters of those RF and DT models were optimized for each prediction task, using a cross-validated technique where the best parameters were chosen and applied in the final model.

\begin{table}[htbp]
\centering
\caption{Area under the ROC curve (AUC) values for the proposed approach and comparison models. The results presented are related to the mean and its 95\% confidence interval.} 
\label{tab:auc}
\begin{tabular}{|l|c|c|c|}
\hline
\textbf{}                    & \textbf{Heart} & \textbf{Breast} & \textbf{Diabetes} \\ \hline
Random forest (no constraints)                & 0.90$\pm$0.01                  & 0.99$\pm$0.00                       & 0.83$\pm$0.01                     \\ \hline
Random forest (simpler)       & 0.87$\pm$0.01                    & 0.98$\pm$0.00                        & 0.79$\pm$0.01                     \\ \hline
Decision tree                & 0.79$\pm$0.02                    & 0.94$\pm$0.01                       & 0.73$\pm$0.01                      \\ \hline
Proposed approach (3 rules)  & 0.82$\pm$0.02                    & 0.97$\pm$0.00                       & 0.70$\pm$0.02                    \\ \hline
Proposed approach (5 rules)  & 0.85$\pm$0.01                   & 0.98$\pm$0.00                       & 0.74$\pm$0.02                  \\ \hline
Proposed approach (10 rules) & 0.89$\pm$0.01                  & 0.99$\pm$0.00                       & 0.79$\pm$0.01                   \\ \hline
Proposed approach (15 rules) & 0.90$\pm$0.01                    & 0.99$\pm$0.00                       & 0.80$\pm$0.01                      \\ \hline
Proposed approach (20 rules) & 0.90$\pm$0.01                    & 0.99$\pm$0.00                    & 0.80$\pm$0.01                     \\ \hline
\end{tabular}
\end{table}

Table \ref{tab:rules} provides additional information, presenting the cross-validation mean number of rules used by the RF and DT methods, in order to compare it with the amount used by our approach (3 to 20 rules).

\begin{table}[htbp]
\centering
\caption{Mean number of rules used by the Random forest and Decision tree models.}
\label{tab:rules}
\begin{tabular}{|l|c|c|c|}
\hline
\textbf{} & \textbf{Heart} & \multicolumn{1}{c|}{\textbf{Breast}} & \multicolumn{1}{c|}{\textbf{Diabetes}} \\ \hline
Random forest (no constraints)                 & \multicolumn{1}{c|}{1230} & 1105 & 969 \\ \hline
Random forest (simpler)       & \multicolumn{1}{c|}{27}  & 29   & 26   \\ \hline
Decision tree                & \multicolumn{1}{c|}{24}  & 18   & 21   \\ \hline
\end{tabular}
\end{table}

\begin{figure}[htbp]
\centering
\includegraphics[width=0.425\textwidth]{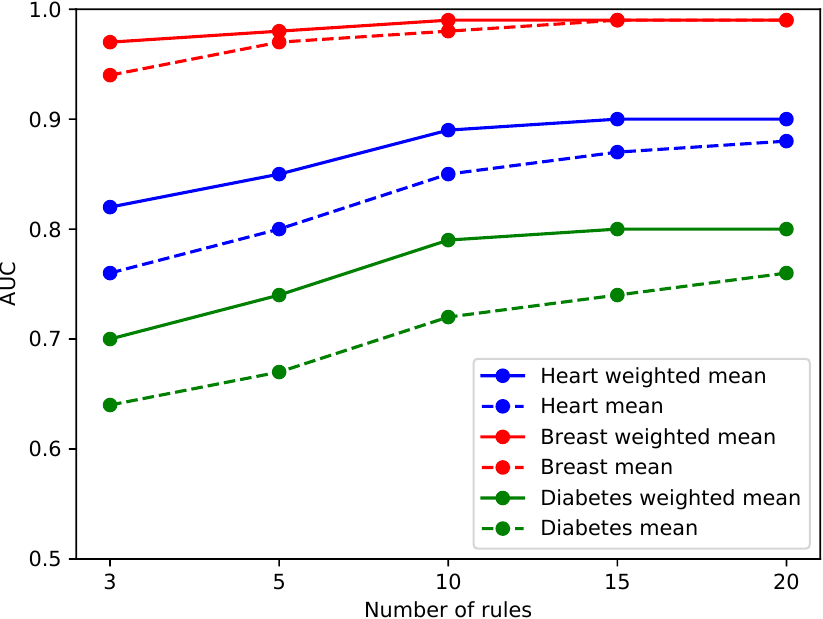}
\caption{Area under the ROC curve (AUC) values for the obtained set of rules, considering a simple mean (standard approaches) and a weighted mean (proposed approach) of their predictions, for the validation datasets.}
\label{fig:meanVSwmean}
\end{figure}

Furthermore, in Fig. \ref{fig:meanVSwmean}, the performance of the models following our approach is presented, i.e. final prediction based on the mean of the rules output weighted by its predicted correctness (2), versus the performance considering only a simple mean of the rules, as in the standard methodologies (1). The weighted mean values are represented by solid lines and the baseline mean values by dashed lines.

Finally, we present an example of a selected subset of decision rules. More specifically, a set for M=3 used in the outcome prediction for the Diabetes dataset:

\begin{enumerate}
  \item IF $nr\_pregnancies \leq 6.5$ AND $[glucose] \leq 124.5$ AND $age \leq 34.5$, THEN no-diabetes (ELSE diabetes).
  \item IF $age \leq 26.5$ AND $BMI \leq 37.2 $ AND $tricep\_skin\_tickness \leq 28.5$, THEN no-diabetes (ELSE diabetes).
  \\
  \item IF $blood\_pressure \leq 69.0$ AND $[glucose] > 119.5$ AND $diabetes\_pedigree \leq 0.23$, THEN diabetes (ELSE no-diabetes).
\end{enumerate}

\section{DISCUSSION}

As we may observe in Table \ref{tab:rules}, the non-constrained RF models (first row) consider a large number of decision rules to generate their outputs. In fact, it would be unfeasible for a physician to analyze thousands of rules in order to interpret the models generated by the RF. Consequently, that is a critical limitation for its application in clinical practice. Further, as presented in Table \ref{tab:auc}, a significant drop in the predictive performance when we use a simpler RF (at most 5 trees with a maximum depth of 3) or a DT instead. 

In Tables \ref{tab:auc} and \ref{tab:rules}, it is possible to observe that the proposed approach achieves better AUC than decision trees using at most 5 rules (while DTs use on average 18 to 24 rules), and equivalent or better AUC than the simpler random forest using at most 10 rules (while simpler RF uses on average 26 to 29 rules). Furthermore, the results show that using only 15 rules, the proposed procedure approaches the AUC performance of the more complex RF models. Therefore, the methodology presented in this paper seems to offer a better compromise between interpretability and prediction performance than the comparison models, which may facilitate its translation to clinical practice.

Fig. \ref{fig:meanVSwmean} shows how the proposed approach compares to the standard methodologies that select rules from ensembles of trees. It is possible to analyze that using the predicted rule's correctness to weigh the final outcome prediction for each patient (giving more importance to the rules predicted to be correct at the individual level) can significantly improve the predictive ability. As expected, this improvement is most noticeable for the smaller sets of rules, as a large set attenuates the effect of a single rule.

Lastly, the prediction of each rule's correctness offers a personalized element in the proposed approach, informing if a given rule is expected to be correct or not for a particular patient. This information can be used by the physicians to further assess the condition of each patient, i.e., they can better evaluate how each rule may be applied individually.

\section{CONCLUSIONS AND FUTURE WORK}

In this study, we introduced an innovative step to the methodologies that aim at extract and select decision rules from ensembles of trees in order to improve its interpretability, while assuring its good prediction performance. Such novelty is related to the prediction of the correctness of each rule for a given patient, which is then used to weigh the final output prediction. This personalization capability seems to improve the ability of the models to correctly classify the patient's outcome. In short, the development of this approach in the clinical domain might assume great importance.

The proposed methodology can be further improved. For example, different methods for rules' subset selection and rules' correctness prediction may be applied, which may increase the performance ability of the models. Finally, its extension to multiclass and regression problems, which are common in the clinical context, may also be considered.

%The methodology proposed in this study can be further improved. For example, different methods for the selection of the rules' subset and for the prediction of rules' correctness may be applied, which may increase the performance ability of the models. Finally, its extension to multiclass and regression problems, which are common in the clinical context, may also be considered

\addtolength{\textheight}{-12cm}   % This command serves to balance the column lengths

%%%%%%%%%%%%%%%%%%%%%%%%%%%%%%%%%%%%%%%%%%%%%%%%%%%%%%%%%%%%%%%%%%%%%%%%%%%%%%%%

%%%%%%%%%%%%%%%%%%%%%%%%%%%%%%%%%%%%%%%%%%%%%%%%%%%%%%%%%%%%%%%%%%%%%%%%%%%%%%%%

%%%%%%%%%%%%%%%%%%%%%%%%%%%%%%%%%%%%%%%%%%%%%%%%%%%%%%%%%%%%%%%%%%%%%%%%%%%%%%%%

\section*{ACKNOWLEDGMENT}

This work was supported by the lookAfterRisk research project (POCI-01-0145-FEDER-030290).

%%%%%%%%%%%%%%%%%%%%%%%%%%%%%%%%%%%%%%%%%%%%%%%%%%%%%%%%%%%%%%%%%%%%%%%%%%%%%%%%

\bibliographystyle{IEEEtran}

\bibliography{references_embc}

\end{document}